\documentclass[10pt]{article}
\usepackage[preprint]{ivgpaper}

\usepackage{graphicx}
\usepackage{booktabs}
\usepackage{multirow}
\usepackage{array}
\usepackage{makecell}
\usepackage{adjustbox}
\usepackage{colortbl}
\usepackage{subcaption}
\usepackage{float}
\usepackage{placeins}
\usepackage{algorithm}
\usepackage{algorithmic}
\usepackage[numbers,sort&compress]{natbib}
\usepackage{url}
\usepackage{tikz}
\usetikzlibrary{arrows.meta,positioning,fit,calc,shapes.geometric,backgrounds}

\newcommand{\tblhead}{\bfseries}
\newcommand{\metric}[1]{\textbf{#1}}

\usepackage{amsmath,amsfonts,bm}

\def\eqref#1{equation~\ref{#1}}
\def\1{\bm{1}}

\DeclareMathAlphabet{\mathsfit}{\encodingdefault}{\sfdefault}{m}{sl}
\SetMathAlphabet{\mathsfit}{bold}{\encodingdefault}{\sfdefault}{bx}{n}

\usepackage[colorlinks=true,linkcolor=ivgblue,citecolor=ivgblue,urlcolor=ivgblue]{hyperref}
\usepackage[capitalize,nameinlink,noabbrev]{cleveref}

\ivgtitle{Syll: Open-Source Personal Automation with Cross-Surface Execution}
\ivgrunningtitle{Syll}
\ivgvenue{Syll Technical Report}
\ivgsetlogo{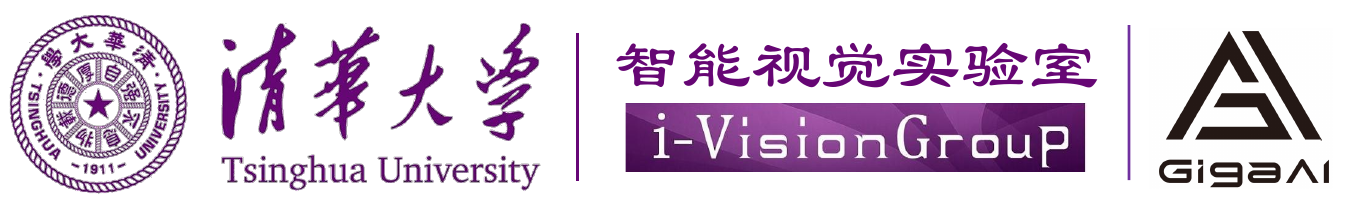}

\ivgauthor{%
  Bo Zhang\textsuperscript{1*}\hspace{0.45em}
  Borui Zhang\textsuperscript{1*{\normalfont$\dagger$}}\hspace{0.45em}
  Chenghao Jiang\textsuperscript{1*}\hspace{0.45em}
  Minglei Shi\textsuperscript{1*}\hspace{0.45em} \\
  \vspace{1mm}
  Xiaofeng Wang\textsuperscript{2}\hspace{0.45em}
  Zheng Zhu\textsuperscript{2}\hspace{0.45em}
  Jie Zhou\textsuperscript{1}\hspace{0.45em}
  Jiwen Lu\textsuperscript{1\#}\\[0.45em]
  \textsuperscript{1}Tsinghua University, \hspace{0.45em} \textsuperscript{2}GigaAI\\[0.25em]
  {\footnotesize
  \textsuperscript{*}Equal contribution. \quad
  \textsuperscript{{\normalfont$\dagger$}}Project leader. \quad
  \textsuperscript{\#}Corresponding author.}\\[0.25em]
  {\footnotesize Project page: \url{https://github.com/THU-SAGE/syll}}
}

\begin{document}
\maketitle

\begin{abstract}
Personal AI agents must increasingly operate across APIs, shells, web surfaces, and desktop GUIs, yet many systems remain tuned to a single interface and offer limited support for user teaching and auditability.
We present \textbf{Syll}, an open-source, self-hosted multimodal agent harness that unifies MCP/API tools, CLI execution, and visual GUI control in a modular runtime, enabling agents to coordinate computer use across heterogeneous interfaces while streamlining how users and agents exchange information.
At the core of Syll is a bidirectional user--agent interaction layer: users teach procedures through direct demonstration, which Syll compiles into reusable skills; agent execution is translated back into multimodal evidence---logs, keyframes, and approval checkpoints---for inspection and control.
Syll further externalizes memory, skills, routines, and governance as editable local artifacts, supporting straightforward inspection, extension, and downstream development. Our implementation has been validated on production desktop applications including Adobe Photoshop, Adobe Audition, Stardew Valley, macOS Finder and others. We report mechanism-oriented studies that validate multimodal routing, teachable GUI replay, and persistent local artifacts. We hope Syll can serve as a practical open-source foundation for personal automation that users can teach, inspect, and continuously extend.
\end{abstract}
\section{Introduction}
\label{sec:introduction}

% 开篇：从真实任务场景引出问题，并强调 system/harness 视角
Personal AI agents are increasingly asked to complete real tasks rather than answer isolated questions. A single request may involve local files, scripts, web dashboards, and closed-source desktop applications whose state is visible only on screen. Strong language modeling is necessary but insufficient. Reliable completion also depends on the \textbf{agent harness}: the layer that constructs context, selects interfaces, coordinates execution, and returns reviewable results. Recent evidence suggests that harness design and agent-computer interface choice materially affect downstream performance \citep{yang2024swe,openai2026harness}.

% 问题一：多模态工作界面，说明为什么单一界面不足
A first unresolved question is how one agent should act across heterogeneous work surfaces. 
\textbf{MCP/API-style tools} are reliable when clean machine interfaces already exist \citep{schick2023toolformer,mcp2026}. 
\textbf{CLI} execution is powerful for composable local control and long-horizon deterministic workflows \citep{yang2024swe}. 
Yet many practical tasks still depend on \textbf{GUI} interaction because state is only visible in pixels, APIs are unavailable, or users require screen-level information \citep{zhou2024webarena,xie2024osworld,qin2025uitars,zhang2026showuialoha}. 
Existing systems often privilege one interface family, forcing a trade-off between reliability and coverage. For personal automation, these surfaces are complementary: each step should use the narrowest interface that can complete the task while preserving the required audit surface.

% 问题二：用户与智能体之间的信息转换，说明 why teachability and auditability matter
A second question is how knowledge should flow between users and agents. 
Many systems assume that operational knowledge must be rewritten as prompts, schemas, or skill files before an agent can use it. This is too restrictive: users may know \emph{how} to perform a task without knowing how to formalize it. Conversely, a raw text trace is often too costly to inspect after a long-horizon task. Personal automation therefore needs a demonstration-to-skill workflow and an evidence pipeline: human know-how becomes reusable skills, and agent execution becomes screenshots, keyframes, logs, diffs, previews, and approval checkpoints.

\begin{figure*}[t]
  \centering
  \includegraphics[width=0.98\textwidth]{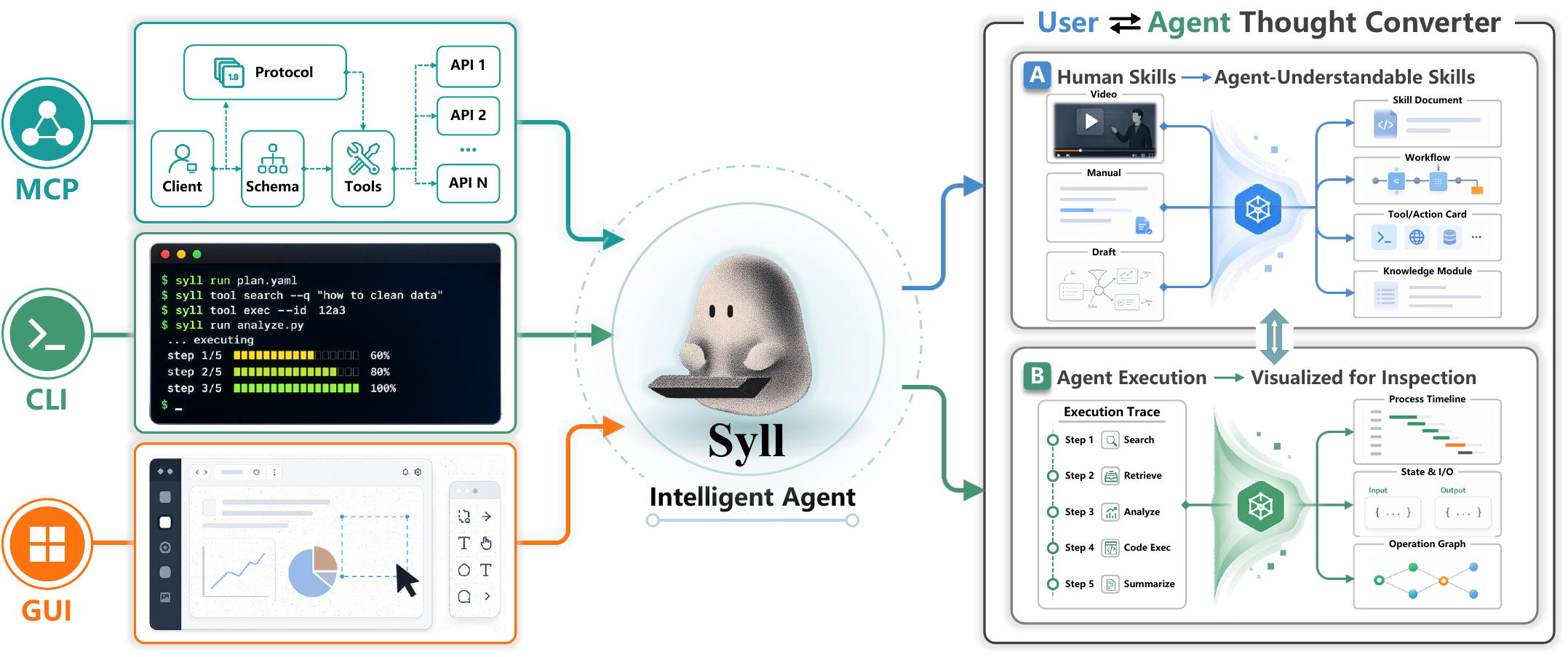}
  \caption{Syll overview. The system unifies three execution surfaces---MCP, CLI, and GUI---around a central multimodal agent harness. The right branch shows the demonstration-to-skill workflow and the audit trail produced from execution evidence.}
  \label{fig:syll_overview}
\end{figure*}

% 方法概述：提出 Syll，并把标题和摘要中的核心术语统一进来
We instantiate this view in \textbf{Syll}, an \textbf{open-source}, self-hosted multimodal agent harness for \textbf{teachable personal automation} shown in \Cref{fig:syll_overview}. 
Syll unifies MCP/API tools, CLI execution, and visual GUI control within one modular runtime. Users teach procedures through direct demonstration, which Syll converts into reusable skills; the runtime translates execution back into multimodal evidence for review. Syll also externalizes memory, skills, routines, governance, and traces as \textbf{persistent local artifacts}, making the system inspectable for users and straightforward to extend for developers.

% 设计意义：强调模块化、开放性与论文后续评估的逻辑
This design is motivated by both usability and research value. 
Keeping execution surfaces, skill conversion, evidence generation, context construction, and confirmation gates explicit reduces cross-component entanglement and lowers the barrier to secondary development. 
It also improves auditability: as task horizons grow \citep{kwameasuring} and open-ended computer-use settings expose verification burden \citep{abhyankar2025osworldhuman}, practical usefulness depends on whether the system is \textbf{teachable}, \textbf{auditable}, and \textbf{extensible}. 
We evaluate these goals through mechanism-oriented studies across production desktop applications including Adobe Photoshop, Adobe Audition, Stardew Valley, and macOS Finder.

% 贡献总结：用和摘要一致的术语收束 introduction
This technical report makes the following contributions:
\begin{itemize}
  \item \textbf{Open-source multimodal agent.} Syll supports MCP/API, CLI, and GUI as complementary execution surfaces and routes tasks to the appropriate layer.
  \item \textbf{Teachable workflow and audit trail.} Direct demonstrations become reusable skills, while execution becomes reviewable evidence.
  \item \textbf{Persistent local artifacts.} Memory, skills, routines, traces, evidence, and governance live as user-editable local artifacts.
  \item \textbf{Mechanism-oriented evaluation.} We validate multimodal routing, teachable GUI replay, and persistent workspace updates.
\end{itemize}

\begin{table}[t]
  \centering
  \footnotesize
  \renewcommand{\arraystretch}{1.12}
  \caption{High-level comparison of agent system families. ``Partial'' means
  the capability exists but is not the primary design center.}
  \label{tab:related_comparison}
  \begin{adjustbox}{width=\linewidth}
  \begin{tabular}{@{}>{\raggedright\arraybackslash}p{0.34\linewidth}cccc>{\raggedright\arraybackslash}p{0.26\linewidth}@{}}
    \toprule
    \tblhead System family & GUI & CLI/API & Demonstration & User memory & Main audience \\
    \midrule
    CLI-only coding agents & No & Yes & No & Partial & developers \\
    MCP/tool frameworks & No & Yes & No & Partial & developers/integrators \\
    GUI  agents & Yes & Partial & Partial & Partial & automation builders \\
    Record-replay/RPA tools & Yes & Partial & Yes & No & operations teams \\
    Companion chat agents & No & Partial & No & Yes & end users \\
    \midrule
    Syll & Yes & Yes & Yes & Yes & users and developers \\
    \bottomrule
  \end{tabular}
  \end{adjustbox}
\end{table}

\section{Related Work}

\subsection{Agent-Centric Task Execution}

Recent progress in autonomous agents has strengthened reasoning, planning, and perception. 
ReAct~\cite{yao2022react} and Reflexion~\cite{shinn2023reflexion} introduced reasoning-action loops for decomposition, environment interaction, and feedback-based refinement. 
Later work extended these ideas to tool invocation and long-horizon execution through external APIs and structured functions~\cite{schick2023toolformer, yang2024swe}. 
% More recently, vision-language models have broadened the interaction surface to graphical user interfaces.
Vision-language models have broadened the interaction surface to graphical user interfaces. 
Systems like Claude Computer Use~\cite{anthropic2026computeruse}, Operator~\cite{openai2026harness}, and UI-TARS~\cite{qin2025uitars} perceive and act upon GUIs through screenshots and low-level control primitives.
Despite these advances, agent-centric approaches usually treat the execution surface as fixed, such as an API, browser, or specific GUI environment, and focus on what the agent decides to do. 
% The resulting interaction knowledge (which pixels to click, which states to wait for, what outcome to expect) remains trapped inside a single trajectory transcript and is rarely externalized as a reusable, auditable artifact that the user or the system can later repurpose.
Consequently, interaction knowledge such as which pixels to click, which states to wait for, and what outcomes to expect often remains in a single trajectory transcript rather than a reusable, auditable artifact.

\subsection{Harness-Centric and Record-and-Replay Systems}

Complementing agent-level work, another line studies infrastructure for reliable, locally owned execution. 
% Complementary to progress in agent intelligence,
Frameworks such as OpenClaw~\cite{openclaw2025} and NanoBot~\cite{nanobot2026} provide local-first runtimes that integrate persistent memory, tool execution, scheduling, and multi-channel communication. 
The Model Context Protocol (MCP)~\cite{mcp2026} standardizes how agents interact with external tools through structured, schema-driven APIs. 
These harness-centric systems are modular and extensible but remain mostly textual or structured; GUI interaction, when present, is often a separate subsystem.

Record-and-replay tools and programming-by-demonstration (PbD) systems let users automate repetitive GUI workflows without scripting. 
Sikuli~\cite{yeh2009sikuli} uses screenshot-driven search to replay visual procedures, CoScripter~\cite{leshed2008coscripter} captures step-by-step action descriptions, and Eager~\cite{cypher1991eager} proactively detects iterative patterns.
More recent efforts like ShowUI-Aloha~\cite{zhang2026showuialoha} incorporate human-taught GUI trajectories for agent learning. 
These approaches capture what the user does, but they often lack semantic phases, a persistent registry shared with non-GUI tools, and an evidence pipeline for inspection. 
As a result, a recorded workflow often remains an isolated macro rather than a runtime abstraction that can be retrieved, scheduled, verified, and refined with other skills.

\subsection{Teachable Automation and Execution Audit}

Two goals central to Syll remain underrepresented: users should be able to \emph{teach} procedures through demonstration, and each executed action should leave \emph{audit evidence}. 
PbD and record-and-replay tools show that demonstration can transfer operational knowledge, but they rarely connect it to a general-purpose agent's context builder or to a shared registry that also contains MCP tools and CLI commands.
Agent evaluation is also largely success-rate driven; benchmarks like OSWorld~\cite{xie2024osworld}, WebArena~\cite{zhou2024webarena}, and OSWorld-Human~\cite{abhyankar2025osworldhuman} measure task completion, while fewer systems structure traces into user-reviewable evidence such as keyframes, diffs, semantic expectations, and approvals.
Syll treats teaching and auditing as one artifact workflow: a demonstration becomes an inspectable skill with visual cues and post-state expectations, while runtime execution over MCP, CLI, or GUI generates a corresponding audit trail.

\subsection{From Fragmented Capabilities to a Unified Artifact Lifecycle}

Taken together, prior work leaves a gap around multi-surface execution, user teaching, and persistent, inspectable evidence in one framework.
Table~\ref{tab:related_comparison} summarizes this landscape: GUI agents provide visual control but limited demonstration handling; MCP/tool frameworks offer structured APIs but little GUI support; record-replay tools support demonstration but omit cross-surface routing and audit; companion chat agents emphasize user memory but not GUI automation.
Syll addresses this gap by making execution surfaces, skill registration, evidence generation, and memory artifacts part of one modular runtime.

\section{Method}
\label{sec:method}

Syll is an open-source, self-hosted computer-use harness for multimodal personal automation. It targets settings where one local agent must preserve context across phone, browser, terminal, local files, scheduled routines, and desktop applications. We model the runtime as an agent loop: the executor runs typed actions through structured tools, shell, or visual GUI control; the context builder supplies the joint external-plus-artifact state; and verification checks evidence, success conditions, and approval gates before commit or retry. Additional details on the runtime, demonstration-skill schema, and formal artifact-option model are provided in the supplementary material.
\label{sec:typed_artifact_option}

\begin{figure}[tbp]
  \centering
  \includegraphics[width=\linewidth]{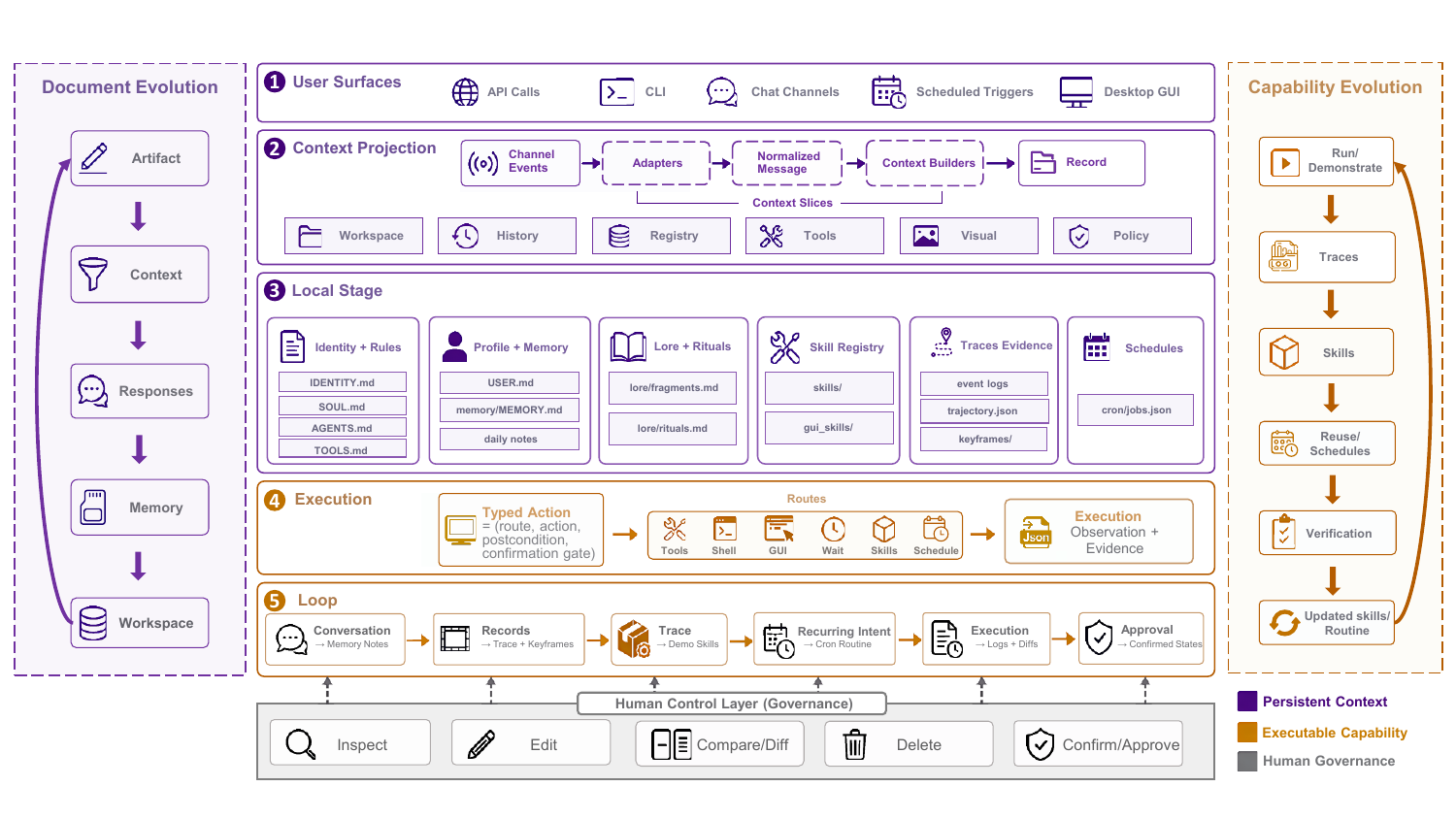}
  \caption{Syll runtime overview. Requests enter one local runtime, are grounded in shared workspace context, routed through tools, commands, or GUI control, and leave audit evidence as confirmations, logs, screenshots, keyframes, and workspace diffs.}
  \label{fig:system_overview}
\end{figure}

\subsection{Multimodal Execution Space}
\label{sec:action_space}
\label{sec:complete_action_space}
\label{sec:design_goals}

Syll targets recurring workflows that span chat, local files, mobile confirmations, and desktop apps. Single-interface agents trade reliability for coverage, so Syll provides three routes and dispatches each request through the narrowest viable one. Schema-driven tools and standardized connectors, including MCP, are preferred when a clean machine interface exists. When the target state is textually accessible, the CLI provides reproducible control through shell, scripts, and files. When state is only visible in pixels, or screen-level evidence is needed, Syll reasons over screenshots and dispatches desktop events. This fallback keeps legacy applications reachable without custom connectors.

\begin{table}[tbp]
  \centering
  \caption{Execution-layer routing policy. Syll selects the narrowest route that
  can complete the task while preserving the audit surface required by policy or
  by the user.}
  \label{tab:action_layer_policy}
  \begin{adjustbox}{width=\linewidth}
  \footnotesize
  \renewcommand{\arraystretch}{1.12}
  \begin{tabular}{@{}>{\raggedright\arraybackslash}p{0.18\linewidth}
    >{\raggedright\arraybackslash}p{0.34\linewidth}
    >{\raggedright\arraybackslash}p{0.35\linewidth}@{}}
    \toprule
    \tblhead Route & Selected when & Evidence retained \\
    \midrule
    Structured tools/API
      & A schema-described operation, connector, file tool, delivery tool, or
        configuration route exists
      & Tool name, validated arguments, output/error, confirmation record \\
    Local command/resource
      & Desired state is machine-readable or reproducible through shell, scripts,
        local files, or browser-accessible resources
      & Command, working directory, stdout/stderr, generated files or diffs \\
    Visual GUI control
      & State is only visible in pixels, no reliable API exists, or the user needs
        screen-level evidence
      & Screenshots, keyframes, low-level actions, visual observations, semantic
        expectations \\
    Confirmation gate
      & External delivery, destructive local change, account change, purchase, or
        scheduled send would create a user-visible side effect
      & Proposal, candidate artifact, approval token, final side-effect record \\
    \bottomrule
  \end{tabular}
  \end{adjustbox}
\end{table}

The route also determines the retained evidence (\Cref{tab:action_layer_policy}). Structured tools keep validated arguments and machine-readable results; the CLI keeps commands, working directories, and outputs; GUI control keeps screenshots, keyframes, and action traces. Side-effectful operations such as file delivery, account changes, scheduled sends, and destructive writes are split into proposal and execution, with commit held until the user supplies an approval token.

\subsection{Demonstration-Teachable Skills and Replay}
\label{sec:demonstration_skills}

Some operational knowledge is easier to demonstrate than to describe. Task phases, visual cues, wait conditions, and expected post-action states are often clearer in a recording than in a prompt. Syll stores each desktop demonstration as a reusable, state-aware skill exposed through the same registry as text-based skills. Building on programming-by-demonstration and demonstration-guided GUI automation \citep{cypher1991eager,leshed2008coscripter,yeh2009sikuli,zhang2026showuialoha}, Syll treats demonstrations as first-class runtime artifacts connected to recording, replay, and audit. Once published, a skill is available through the web UI, natural language invocation, and scheduled routines.

\begin{figure}[tbp]
  \centering
  \begin{subfigure}{\linewidth}
    \centering
    \includegraphics[width=0.98\linewidth, trim={0 198 0 0}, clip]{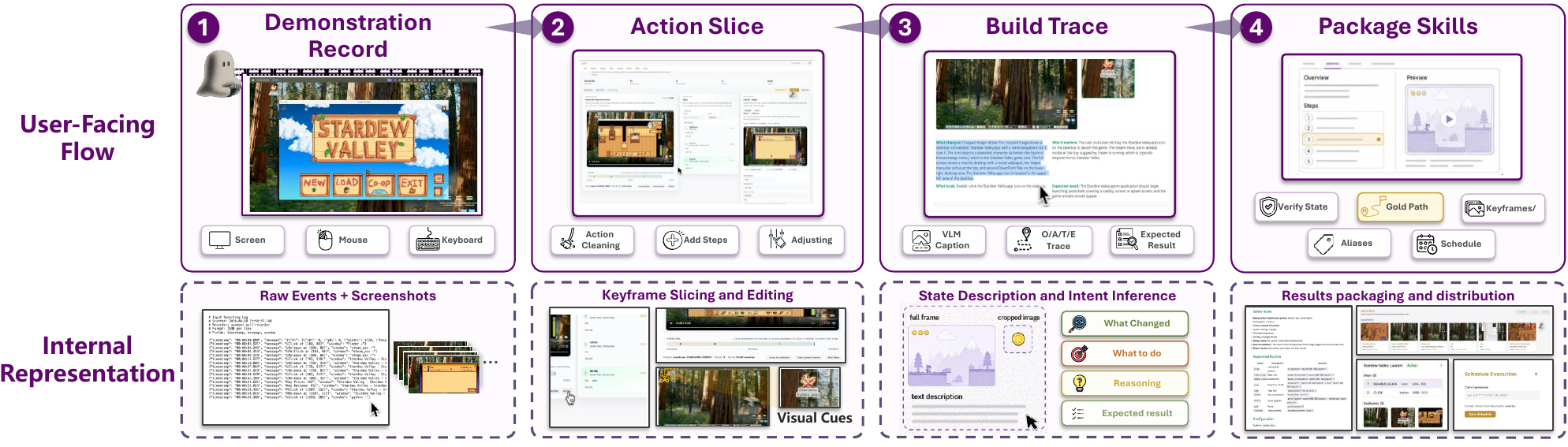}
    \caption{Demonstration recording and packaging. The four user-facing steps (\emph{Record}, \emph{Action Slice}, \emph{Build Trace}, \emph{Package Skills}) turn raw screen, mouse, and keyboard events into a reusable skill directory. The internal row below shows the on-disk artifact each step produces.}
    \label{fig:demo_recording}
  \end{subfigure}\\[0.4em]
  \begin{subfigure}{\linewidth}
    \centering
    \includegraphics[width=0.98\linewidth, trim={0 0 0 272}, clip]{figures/syll_GUI.pdf}
    \caption{State-aware replay. \emph{Trajectory learning} reuses the recorded demonstration as a reference path and an expectation source, emitting \textsc{Advance}, \textsc{Wait}, or repair signals to the runtime guide. \emph{Action ICL} grounds each next GUI action against demonstrated long-tail UI cues, falling back to \textsc{Recover} when no target is grounded.}
    \label{fig:demo_replay_loop}
  \end{subfigure}
  \caption{Demonstration-teachable skills in Syll. (a) Recording and packaging. (b) State-aware replay.}
  \label{fig:demo_replay}
\end{figure}

Replay combines phase indexing with in-context action grounding (Action ICL), which packs the previous, current, and next demonstrated steps into the actor's window (\Cref{fig:demo_replay_loop}). The phase index tracks the active demonstration step, while the packed window grounds actions against local UI cues such as labels, repeated controls, and similar icons. After each action, the actor captures a screenshot and verifies it against current and next expectations. The index advances when either expectation matches; otherwise it holds for recovery. \textsc{Wait} marks transient states such as loading screens and launches, and \textsc{Recover} re-grounds without advancing the index. This preserves state across asynchronous transitions where a screenshot-only policy may repeat completed phases. The supplementary material gives the full replay episode.

\subsection{Document- and Capability-Evolution}
\label{sec:document_evolution}
\label{sec:capability_evolution}

Syll updates persistent context and executable capabilities rather than model weights. Identity, memory, routines, skills, traces, and evidence live as editable files refined by the user and by Syll's own writes; the supplementary material lists the artifact categories and filenames. During context construction, the builder resolves workspace variables and injects relevant document slices into the prompt, so user or agent edits affect later behavior. The resulting state is a set of files the user can back up, audit, or delete.

Two loops sit on top of this workspace. The \emph{Document Evolution} loop lets users refine identity, rules, profile, lore, and routine files in plain text while Syll updates memory artifacts from conversations, tool outcomes, and feedback. The \emph{Capability Evolution} loop turns runtime interactions into artifacts the agent can retrieve, execute, inspect, and refine. Traces store screenshots, tool arguments, and confirmations; skills distill repeatable procedures; routines attach skills to recurring triggers. Habits therefore accumulate as auditable automation rather than hidden model state.

Per-layer evidence and approval gates form Syll's trust boundary: past actions are reconstructible, and side-effectful next steps require confirmation before commit. Channels, tools, actors, schedules, and skill types use explicit extension boundaries, so researchers can extend the runtime without rewriting the core loop.

The name \emph{Syll} encapsulates this stance. A syllable means little on its own and acquires meaning from the surrounding utterance, and the agent likewise begins with a structured but incomplete identity that gains specificity from accumulated documents.

\section{Evaluation}
\label{sec:experiments}

Our evaluation targets heterogeneous personal automation across chat, local files, CLI, browser sessions, desktop GUI, and scheduled triggers. We use three mechanism-oriented studies. \textbf{Study~A} probes execution-route selection when the task does not specify an interface. \textbf{Study~B} isolates the contribution of recorded demonstration to GUI replay. \textbf{Study~C} follows one user across five simulated-month episodes and tracks workspace updates. Each number comes from a run log or deterministic checker output. Timing, model calls, tokens, screenshots, and estimated cost are reported as local efficiency metadata, not cross-system rankings. \Cref{tab:eval_matrix} summarizes the contrasts and readouts.

\begin{table}[tbp]
  \centering
  \caption{Evaluation map. Each study isolates one mechanism and reports a compact auditable readout.}
  \label{tab:eval_matrix}
  \footnotesize
  \renewcommand{\arraystretch}{1.18}
  \setlength{\tabcolsep}{4pt}
  \begin{tabular}{@{}>{\raggedright\arraybackslash}p{0.17\linewidth}
    >{\raggedright\arraybackslash}p{0.25\linewidth}
    >{\raggedright\arraybackslash}p{0.28\linewidth}
    >{\raggedright\arraybackslash}p{0.24\linewidth}@{}}
    \toprule
    \tblhead Study & Mechanism isolated & Controlled contrast & Main readout \\
    \midrule
    \makecell[l]{\metric{Study A}\\Execution routing}
      & route choice across structured tools, shell, and visual GUI control
      & file delivery, local command, and visual GUI sub-probes (inspection plus two creative-app actuations)
      & \metric{3/3} routes matched and trajectories completed with audit artifacts retained \\
    \makecell[l]{\metric{Study B}\\Demo replay}
      & effect of a recorded trajectory in clean GUI replay
      & demo-guided replay vs instruction-only replay
      & \metric{3/3} vs \textbf{2/3} phases; \metric{0} vs \textbf{3} off-trace actions \\
    \makecell[l]{\metric{Study C}\\Workspace updates}
      & persistence, refinement, and routine promotion
      & Full Syll vs Frozen Workspace across five delayed episodes
      & \metric{5/5 + 4/4} reuse; \metric{refine=1}; \metric{routine=1} \\
    \bottomrule
  \end{tabular}
\end{table}

The studies are complementary. Studies~A and~B exercise the runtime under model-driven probes with a shared model, executor, and harness. Study~C uses a deterministic artifact checker driven by scripted user turns, so its outcomes are independent of how well the model extracts preferences. Aggregate run metadata for the two model-driven studies appear in \Cref{tab:efficiency_template}.

\subsection{Study A on Cross-Surface and Execution-Space Coverage}
\label{sec:action_space_eval}

Study~A tests whether Syll chooses the appropriate execution layer when the task does not name an interface. A run passes when the selected layer matches the intended interface, the trajectory completes, audit evidence is retained, and no side-effect error occurs. Timing, calls, tokens, and screenshots are reported but not used for pass/fail. \textbf{A1} finds a local presentation by visual attribute and delivers it through a side-effectful channel, testing structured file tools and confirmation. \textbf{A2} tests command-layer routing when a single shell call exposes the requested state. \textbf{A3} tests Visual GUI control: \textbf{A3a} is screen-only inspection, while \textbf{A3b} (Photoshop on a $1024{\times}1024$ image) and \textbf{A3c} (Audition on a 6.2~s speech clip) require multi-step GUI actuation.

\begin{table}[tbp]
  \centering
  \caption{Study A result ledger. A3 has three sub-tasks (A3a inspection,
  A3b Photoshop, A3c Audition). \texttt{---} in Time/Calls/Tokens marks
  rows where GUI step counts replace token-level metadata.}
  \label{tab:e1_template}
  \footnotesize
  \renewcommand{\arraystretch}{1.06}
  \setlength{\tabcolsep}{3pt}
  \begin{adjustbox}{max width=\linewidth,center}
  \begin{tabular}{@{}>{\raggedright\arraybackslash}p{0.15\linewidth}
    >{\raggedright\arraybackslash}p{0.13\linewidth}
    >{\raggedright\arraybackslash}p{0.27\linewidth}
    >{\raggedright\arraybackslash}p{0.18\linewidth}
    >{\raggedleft\arraybackslash}p{0.06\linewidth}
    >{\raggedleft\arraybackslash}p{0.05\linewidth}
    >{\raggedleft\arraybackslash}p{0.08\linewidth}@{}}
    \toprule
    \tblhead Probe & Route & Trajectory outcome & Evidence & Time & Calls & Tokens \\
    \midrule
    \metric{A1 File}
      & structured $\rightarrow$ confirm
      & attached the green Aurora roadmap deck
      & previews, paths, confirmation
      & 25.0\,s & 6 & 85.7\,k \\
    \metric{A2 CLI}
      & local CLI
      & returned \texttt{cwd} and top-level entry count
      & command log, stdout
      & 4.5\,s & 2 & 26.6\,k \\
    \cmidrule(lr){1-7}
    \multicolumn{7}{@{}l}{\metric{A3 Visual GUI control}} \\
    \quad A3a~Inspect
      & visual screenshot
      & captured one screenshot and summarized visible UI state
      & screenshot, state summary
      & 13.3\,s & 2 & 28.9\,k \\
    \quad A3b~Photoshop
      & GUI on Photoshop
      & drove PS cutout in 5 GUI steps, exporting alpha PNG and editable PSD
      & screenshots, PSD, alpha PNG, report
      & --- & --- & --- \\
    \quad A3c~Audition
      & GUI on Audition
      & drove Audition voice repair in 17 GUI steps, exporting cleaned WAV
      & screenshots, cleaned WAV, report
      & --- & --- & --- \\
    \midrule
    \metric{Suite}
      & \textbf{3/3 match}
      & \textbf{5/5 trajectories completed}
      & \textbf{1 confirm gate, 1 inspection, 2 artifacts}
      & \textbf{42.8\,s} & \textbf{10} & \textbf{141.1\,k} \\
    \bottomrule
  \end{tabular}
  \end{adjustbox}
\end{table}

The ledger shows three route choices without manual routing hints, matching the narrowest-viable-route policy in \Cref{tab:action_layer_policy}. Evidence also adapts to the layer: A1 returns previews and a confirmation token, A2 returns a command log and stdout, and A3 returns screenshots plus either a state summary or exported artifacts. A3 covers both Visual GUI triggers, with A3a as inspection and A3b/A3c as actuation. The same GUI loop drives Photoshop raster editing and Audition audio editing without per-application code paths, motivating Study~B's closer look at GUI actuation.

\subsection{Study B on Demonstration-Trajectory Ablation}
\label{sec:demobench}

Study~B isolates the recorded demonstration. We fix the GUI actor, executor, task instruction, and initial setup, varying only whether Syll provides the demonstration skill artifact. The contrast probes both Action ICL grounding for local UI cues and demonstration-conditioned phase indexing across asynchronous transitions.

The task launches Stardew Valley from the macOS desktop, opens the Load menu, and selects a specific farm save (\texttt{Vlm}/\texttt{VLMBot}). It stresses replay information that is hard to recover from the current screenshot alone: the Load button, an hourglass save icon, and the player-and-farm label. The agent must pass launch, load-menu, and farm phases in order, while \textsc{Wait}/\textsc{Recover} handles loading and animation states. The farm-world HUD gives a deterministic success condition.

In \textbf{demo-guided replay}, the actor receives the instruction $u$, screenshot $I_k$, and the demonstration artifact with keyframes $\mathit{kf}$ and semantic traces $\mathit{trace}$. \textbf{Instruction-only replay} withholds the artifact, so the actor sees only $u$ and $I_k$. Differences in phase coverage, off-trace actions, or cue grounding are therefore attributable to the demonstration. The supplementary material lists the skill schema.

\begin{table}[tbp]
  \centering
  \caption{Study B clean demonstration-trajectory ablation. Demonstration
  guidance mainly changes phase indexing and long-tail cue grounding, not the
  model-call budget.}
  \label{tab:e2_template}
  \footnotesize
  \renewcommand{\arraystretch}{1.13}
  \setlength{\tabcolsep}{3pt}
  \begin{tabular}{@{}>{\raggedright\arraybackslash}p{0.20\linewidth}
    >{\raggedright\arraybackslash}p{0.39\linewidth}
    >{\raggedright\arraybackslash}p{0.35\linewidth}@{}}
    \toprule
    \tblhead Evidence item & Demo-guided replay & Instruction-only replay \\
    \midrule
    Result
      & \metric{Pass}; farm HUD visible, Wed.~3, 6:00\,am, 50g
      & \textbf{Fail}; target save not loaded \\
    Phase coverage
      & \metric{3/3}; launch $\rightarrow$ Load $\rightarrow$ farm
      & \textbf{2/3}; save phase missing \\
    Action accounting
      & \metric{3 semantic actions}; 8 state checks; \metric{0 off-trace}
      & \textbf{15 GUI steps}; \textbf{3 off-trace} \\
    Phase indexing
      & waits through launch and menu loading; \metric{0 order violations}
      & \textbf{10 repeated phase visits}; revisits earlier phases after state changes \\
    Cue grounding
      & \metric{4/4}; app icon, Load button, save identity, hourglass
      & \textbf{2/4}; app icon and Load button only \\
    Efficiency anchors
      & \metric{148.2\,s}; 8 shots; 16 calls; 43.3\,k tokens
      & \textbf{201.1\,s}; 15 shots; 17 calls; 43.2\,k tokens \\
    \bottomrule
  \end{tabular}
\end{table}

Instruction-only replay reproduces the failure mode in \Cref{sec:demonstration_skills}. The actor grounded the app icon and Load button, but exhausted the 15-step budget without selecting the \texttt{Vlm}/\texttt{VLMBot} save and revisited completed phases ten times. The failure lies in phase indexing and post-action checking rather than low-level GUI actuation.

Demo-guided replay reaches the final farm state with similar model-call and token totals (16 vs 17 calls, 43.3\,k vs 43.2\,k tokens). Action ICL grounds all four long-tail cues, including the hourglass icon and player-and-farm label. Demonstration-conditioned phase indexing keeps the actor on path with zero off-trace actions and zero phase regressions, while \textsc{Wait} absorbs launch, menu, and save-list transitions. The 52.9\,s wall-time and 46.7\% screenshot reductions follow from these effects rather than from an explicit efficiency objective.

\subsection{Study C on Month-Scale Workspace Updates}
\label{sec:coevolution_pilot}

Study~C evaluates persistent workspace updates as inspectable artifact changes. The scenario follows a weekly project review over a simulated month: the user teaches formatting preferences and writing style, reuses them, refines one preference, and promotes the workflow into a scheduled routine. We compare Full Syll, which persists artifacts across episodes, against a Frozen Workspace baseline that resets them. A deterministic artifact checker executes the sequence; credit is given only for before/after diffs, approval records, reuse hits, or routine artifacts.

The contrast is between \textbf{Full Syll}, where \texttt{USER.md}, \texttt{SOUL.md}, and routine artifacts persist across episodes, and \textbf{Frozen Workspace}, where each episode performs the same per-episode extraction but resets artifacts before the next episode begins. The two conditions differ only in artifact persistence, so any difference in later reuse, refinement, or routine promotion is attributable to it.

\begin{table}[tbp]
  \centering
  \caption{Study C month-scale workspace update ledger. Metric styling highlights the artifact transition.}
  \label{tab:e3_template}
  \footnotesize
  \renewcommand{\arraystretch}{1.16}
  \setlength{\tabcolsep}{3pt}
  \begin{tabular}{@{}>{\raggedright\arraybackslash}p{0.10\linewidth}
    >{\raggedright\arraybackslash}p{0.16\linewidth}
    >{\raggedright\arraybackslash}p{0.12\linewidth}
    >{\raggedright\arraybackslash}p{0.33\linewidth}
    >{\raggedright\arraybackslash}p{0.23\linewidth}@{}}
    \toprule
    \tblhead Time & Episode & Artifact & Full Syll & Frozen Workspace \\
    \midrule
    \makecell[l]{Week 1\\Day 1} & Preference capture & \texttt{USER.md}
      & \metric{+5 USER facts}; Aurora source, format, length, delivery preference; diff saved
      & captures current-turn facts; reset before next episode \\
    \makecell[l]{Week 1\\Day 3} & Style alignment & \texttt{SOUL.md}
      & \metric{+4 SOUL rules}; calm technical voice approved; diff saved
      & accepts current-turn style patch; reset before next episode \\
    \makecell[l]{Week 2\\Day 8} & Joint reuse & both
      & \metric{reuse 5/5 + 4/4}; 0 clarifications; 39 user words saved
      & 0/5 facts + 0/4 rules retained; 1 clarification \\
    \makecell[l]{Week 3\\Day 15} & Preference refinement & \texttt{USER.md}
      & \metric{refine=1}; Blockers made explicit; USER diff saved
      & missing prior context; 1 clarification; no refinement diff \\
    \makecell[l]{Week 5\\Day 29} & Routine promotion & routine
      & \metric{routine=1}; refined Blockers, voice policy, confirmation retained
      & no qualified routine; 1 clarification \\
    \bottomrule
  \end{tabular}
\end{table}

Both conditions capture preferences and style rules within one turn, so one-shot extraction does not require persistence. They diverge from Week~2 onward. Only Full Syll reuses captured artifacts for the short request ``Do my Aurora weekly review.'' A week later, only Full Syll has the prior context needed to produce an auditable \texttt{USER.md} diff for the refined Blockers preference. By Week~5, only Full Syll promotes a qualified \texttt{aurora\_weekly\_review} routine referencing the Aurora source file, three-section format, refined Blockers preference, \texttt{Work-Report Voice} rule, and draft-first confirmation. The contrast shows that artifact persistence enables later reuse, refinement, and routine promotion.

\subsection{Discussion}
\label{sec:eval_discussion}

\paragraph{Scope and limitations.} Each study has a narrow
scope. The route probes in \textbf{Study~A} are hand-constructed on one machine, so they validate the selected routes but say nothing about adversarial routing coverage. \textbf{Study~B} is a single paired replay ($n=1$ per condition) that exercises the demonstration schema and replay loop, not statistical robustness or environmental perturbation (resolution, theme, language). \textbf{Study~C} uses a deterministic checker with scripted user turns, so it verifies workspace and routine behavior rather than model extraction quality, population-level preference modeling, or long-term retention. Broader task sets, perturbation studies, and human audit studies are left for future work.

\paragraph{Reproducibility.} The release package includes task setup
scripts, Study~C's deterministic driver, success checks, run-date prompt and provider configuration, workspace layout, JSONL event logs, GUI replay evidence, and the pricing table used for cost fields. API keys, channel credentials, and private user data are excluded. \Cref{tab:efficiency_template} reports aggregate metadata for Studies~A and~B, and \Cref{tab:e1_template} and \Cref{tab:e2_template} give the per-run breakdowns.

\begin{table}[tbp]
  \centering
  \footnotesize
  \renewcommand{\arraystretch}{1.12}
  \caption{Aggregate run metadata for the model-driven studies A and B.
  Study~A's row covers the token-eligible probes A1, A2, and A3a (visual
  inspection). The creative-app actuation sub-probes A3b and A3c are reported
  in \Cref{tab:e1_template} with GUI step counts in place of token-level
  metadata, since their GUI loops are dominated by visual perception rather
  than language generation. Study~C is omitted because its driver is
  deterministic and not a model-efficiency anchor.}
  \label{tab:efficiency_template}
  \begin{tabular}{@{}lrrrrr@{}}
    \toprule
    \tblhead Aggregate & Wall time & Calls & Input tokens & Output tokens & Shots \\
    \midrule
    \metric{A-suite (A1, A2, A3a)}   & 42.8\,s  & 10 & 140.3\,k & 0.8\,k &  4 \\
    \metric{B-pair (2 runs)}    & 349.3\,s & 33 &  86.0\,k & 0.5\,k & 23 \\
    \bottomrule
  \end{tabular}
\end{table}

\section{Conclusion}
\label{sec:conclusion}

% 总结问题与核心 insight：回扣标题、摘要与全文主线
This paper presents Syll as an \textbf{open-source multimodal agent harness} for \textbf{teachable personal automation}. Practical computer use is not only a model problem; it also requires routing work across MCP/API tools, CLI execution, and GUI control, while turning human procedures into reusable skills and agent behavior into reviewable evidence.

% 总结系统设计与主要发现：强调 teachable / auditable / extensible
Syll implements this view through a self-hosted runtime with persistent local artifacts for memory, skills, routines, traces, approvals, and governance. Our mechanism-oriented studies validate multimodal routing, teachable GUI replay, and persistent workspace updates. Together, the results suggest that open-source agent systems become more usable when \textbf{teachability}, \textbf{auditability}, and \textbf{extensibility} are treated as core system objectives.

% 局限与展望：保持克制，说明 scope，并给出下一步方向
Syll still has scope limitations. The current evaluation validates core mechanisms rather than large-scale benchmark superiority across operating systems, interface distributions, or user populations. Next steps include broader task suites, stronger visual grounding verification, longer-horizon memory consolidation, richer MCP/API integrations, and more robust policies for side-effectful actions.

\bibliographystyle{plainnat}
\bibliography{references}

\end{document}